\documentclass[a4paper, 11pt]{article}
\usepackage{graphicx} % Required for inserting images
\usepackage{fancyhdr}
\usepackage{times}
\usepackage{amsfonts}
\usepackage{subcaption}
\usepackage{graphicx}
\usepackage{adjustbox}
\usepackage{booktabs}
\usepackage[dvipsnames]{xcolor}
\usepackage{times}
\usepackage{linguex}
\usepackage{inconsolata}
\usepackage[margin=1in]{geometry}
\usepackage[numbers, sort, compress]{natbib}
\usepackage[colorlinks = true,
           linkcolor = black,
           urlcolor  = BrickRed,
           citecolor = BrickRed,
           anchorcolor = blue]{hyperref}

\newcommand{\kimsmol}{K\&S}
\newcommand{\fitb}{\textsc{fitb}}
\newcommand{\noun}{\textsc{noun}}
\newcommand{\vrb}{\textsc{verb}}
\newcommand{\adj}{\textsc{adj}}
\newcommand{\adv}{\textsc{adverb}}
\newcommand{\blank}{$\rule{0.6cm}{0.15mm}$}

\begin{document}

\begin{center}
    \large\textbf{Abstraction via exemplars? A representational case study on lexical category inference in BERT}
    \\
    Kanishka Misra$^{\tau, \pi}$ and Najoung Kim$^\beta$\\
    $^\tau$The University of Texas at Austin, $^\pi$Purdue University, $^\beta$Boston University
\end{center}

\noindent
Linguistic abstractions are often posited as the mechanism by which language learners generalize to novel expressions. However, \citet{ambridge2020against} argues that such abilities arise not through linguistic abstraction but instead via stored exemplars encountered during learning. One counterargument \citep{mahowald2020counts}, among many \citep{ambridge2020abstractions}, takes inspiration from the unprecedented success of neural network language models (LMs) such as BERT \citep{devlin-etal-2019-bert}.
Specifically, LMs deviate from both the radical exemplar account and the pure abstraction-based account---their parameters store a summarized form of their training exemplars as guided by their training objective (word-prediction in context), rather than storing individual exemplars or encoding abstractions explicitly. Empirically, this has shown to give rise to generalization behaviors aligning with abstract linguistic structures \citep{goldberg2019assessing}. Abstraction could therefore be viewed as arising \textit{via} the (compressed) encoding of exemplars.

\begin{figure}[t!]
    \centering
    \includegraphics[width=0.7\textwidth]{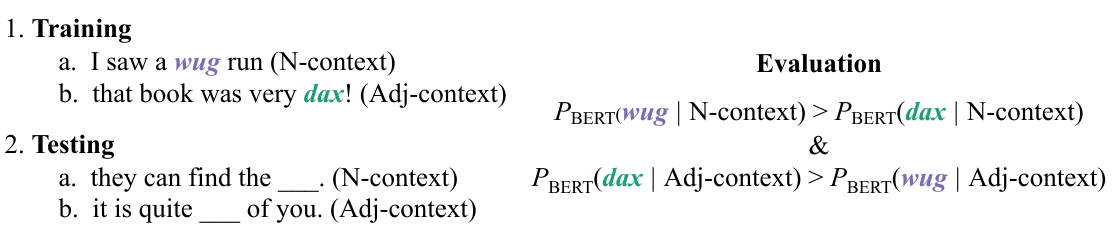}
    \caption{
    Experimental setting proposed by \citet{kim2021testing}, illustrated with \noun{} vs. \adj{}.
    }
    \label{fig:kimsmol1}
\end{figure}

\begin{figure}[!t]
    \centering
    \includegraphics[width=0.65\textwidth]{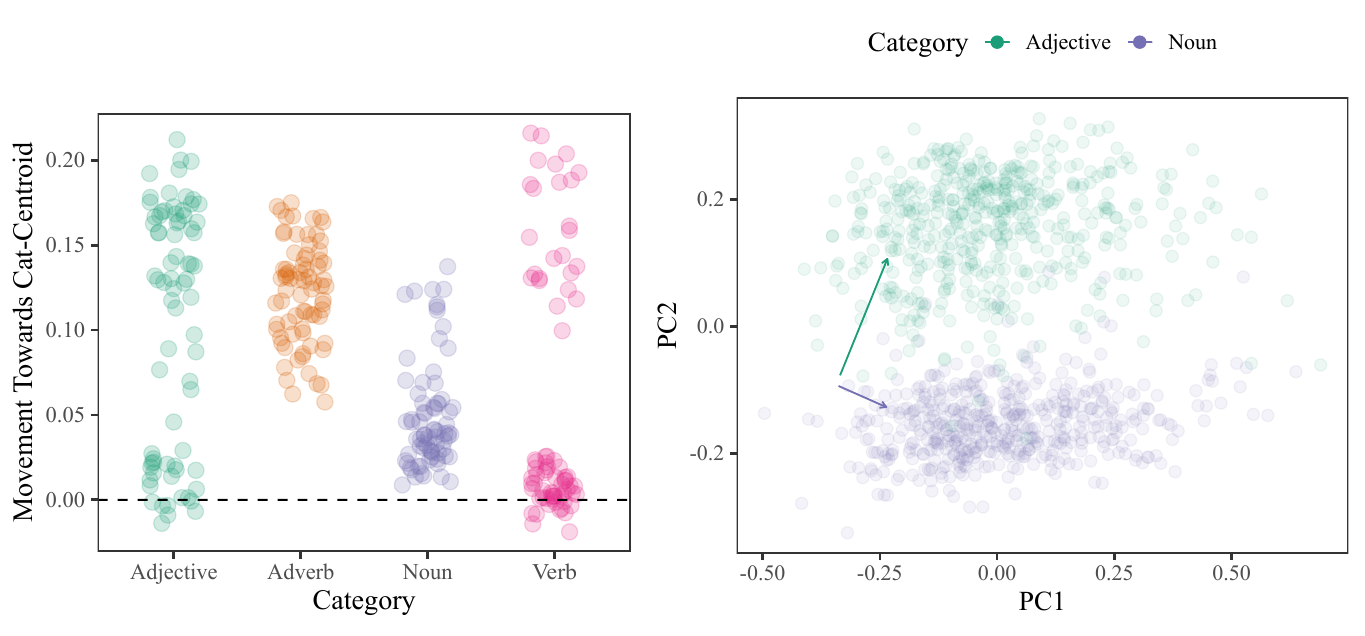}
    \caption{
    \textbf{Left:} Relative movement of the novel token representations with respect to known category exemplars for each category after training on the \kimsmol{} experiments. 0.0 indicates no movement. \textbf{Right:} Movement (indicated by arrow) of a novel token's two-dimensional representation from its initial state in the \adj{}--\noun{} experiment. Points indicate known, unambiguous adjectives and nouns.
    }
    \label{fig:movement}
\end{figure}

In this work, we provide further empirical evidence for the \textit{abstraction-via-exemplars} account, by presenting a case study on lexical category inference in BERT, an LM that is trained to predict missing words in fill-in-the-blank (\fitb) inputs (e.g., \textit{I \blank{} you. $\rightarrow$ ``like''}). 
We adapt the experiments of \citet[hereafter \kimsmol{}]{kim2021testing}, who tested BERT's generalization of the usage of novel tokens belonging to either \noun{}/\vrb{}/\adj{}/\adv{}, from exposure to a single observation
Their tests are inspired by developmental linguistic studies that probe for lexical category abstraction in children \citep{hohle2004functional}.
Briefly, \kimsmol{} used pairs (see Figure \ref{fig:kimsmol1}) of \fitb{} stimuli that disambiguate the category of the novel token and trained the model's embeddings of the novel tokens only (initialized with random-values), keeping all other parameters constant. 
They then tested BERT's accuracy for predicting each of the novel tokens on a set of \fitb{} inputs (N=100 per category) that are lexically disjoint from the training examples, and found overall above-chance performance indicating desirable generalization patterns.
Using representations learned by this method, we perform two analyses that relate the representational change in the novel tokens' embeddings with BERT's generalization patterns on the \kimsmol{} experiment.
First, we track the movement of the embeddings in two-dimensional space (obtained using Principal Component Analysis) as they are updated during training. 
Figure \ref{fig:movement} suggests that the final states of the embeddings of the novel tokens move closer in two-dimensional space to that of known, unambiguous category exemplars (N=500 per category).
That is, regardless of initialization, \textit{wug} (Figure \ref{fig:kimsmol1}) becomes more noun-like, while \textit{dax} becomes more adjective-like.
To further analyze this behavior, we take multiple novel tokens (N=20 per category) and assign them values randomly sampled from regions in two-dimensional space populated by category exemplars.
We then project them into BERT's embedding space, yielding novel token representations with values informed by two-dimensional encoding of category regions. 
Evaluating BERT to predict these novel tokens on the \kimsmol{} testing set---without any additional training---yields substantially above-chance performance across multiple category pairs (Table \ref{tab:pcaemb}).

Together, these analyses suggest that representational movement towards regions of known category exemplars can result in behavior that is consistent with successful generalization to novel expressions.
Overall, this lends empirical support to the claim that abstraction and exemplar based accounts need not be at odds, and that abstraction-like behavior can emerge in learners that encode training exemplars.
\\

\noindent
\hspace*{0pt}\hfill (498 words)

\newpage

\begin{table}[!t]
\centering
\resizebox{0.3\columnwidth}{!}{%
\begin{tabular}{@{}lc@{}}
\toprule
\textbf{Category Pair} & \textbf{Accuracy} \\ \midrule
% \textsc{Chance Level} & $0.50$\\
\adj--\adv & $0.93_{\pm0.03}$ \\
\adj--\vrb & $0.70_{\pm0.06}$ \\
\adv--\vrb & $0.87_{\pm0.05}$ \\
\noun--\adj & $0.80_{\pm0.08}$ \\
\noun--\adv & $0.89_{\pm0.04}$ \\
\noun--\vrb & $0.81_{\pm0.08}$ \\ \bottomrule
\end{tabular}%
}
\caption{
Accuracies (with 95\% CI) on the test set of \kimsmol{} \citep{kim2021testing} obtained by randomly sampling values from two-dimensional regions of category-exemplars (e.g., see Fig 1-Right) which are projected to serve as BERT embeddings for novel, unseen tokens (N=20 each). \textbf{Chance performance is 0.50}.
}
\label{tab:pcaemb}
% \vspace{-1em}
\end{table}

\scriptsize
\bibliographystyle{unsrtnat}
\bibliography{refs}

\end{document}